\documentclass[conference, a4paper]{IEEEtran}
\usepackage[
  left=54pt,
  right=37pt,
  bottom=54pt,
  top=104pt
]{geometry}
\IEEEoverridecommandlockouts
\usepackage{cite}
\usepackage{amsmath,amssymb,amsfonts}
\usepackage{algorithmic}
\usepackage{graphicx}
\usepackage{textcomp}
\usepackage{xcolor}
\usepackage{listings}
\usepackage{hyperref}
\hypersetup{
  pdftitle={Hierarchical Prompting with Dual LLM Modules for Robotic Task and Motion Planning},
  pdfauthor={Karolina Źróbek; Tessa Pulli; Paweł Gajewski; Antonio Galiza Cerdeira Gonzalez; Bipin Indurkhya},
  pdfsubject={Robotic task and motion planning with large language models},
  pdfkeywords={Large Language Models, Robotic Agents, Spatial Reasoning, Task Planning}
}
\usepackage{booktabs}
\usepackage{url}
\def\BibTeX{{\rm B\kern-.05em{\sc i\kern-.025em b}\kern-.08em
    T\kern-.1667em\lower.7ex\hbox{E}\kern-.125emX}}

\usepackage[utf8]{inputenc}
\usepackage[most]{tcolorbox}
\usepackage{makecell}

\definecolor{qgreen}{RGB}{100, 180, 100}   
\definecolor{tyellow}{RGB}{218, 165, 32}    
\definecolor{apink}{RGB}{240, 128, 172} 
\definecolor{ipurple}{RGB}{180, 130, 200} 
\definecolor{obsred}{RGB}{255, 120, 100} 
\definecolor{fanswer}{RGB}{200, 60, 60}

\newtcolorbox{planbox}{
    enhanced,
    colback=gray!2,
    colframe=gray!30,
    arc=1.5mm,
    boxrule=0.5pt,
    left=15pt,
    right=15pt,
    top=12pt,
    bottom=12pt
}

\newcommand{\thought}[1]{\noindent\textbf{\textcolor{tyellow}{Thought:}} \textit{#1}\par\smallskip}
\newcommand{\action}[1]{\noindent\textbf{\textcolor{apink}{Action:}} \texttt{#1}\par\smallskip}
\newcommand{\inputval}[1]{\noindent\textbf{\textcolor{ipurple}{Action Input:}} \texttt{#1}\par\smallskip}
\newcommand{\observation}[1]{%
    \noindent\textbf{\textcolor{obsred}{Observation:}} #1\par\medskip
}

\begin{document}

\title{\vspace*{50pt}Hierarchical Prompting with Dual LLM Modules for Robotic Task and Motion Planning}



\author{
\IEEEauthorblockN{
Karolina Źróbek\textsuperscript{1},
Tessa Pulli\textsuperscript{2},
Paweł Gajewski\textsuperscript{3,4},
Antonio Galiza Cerdeira Gonzalez\textsuperscript{3},
Bipin Indurkhya\textsuperscript{3}
}
\IEEEauthorblockA{\textsuperscript{1}IBM, Krakow, Poland}
\IEEEauthorblockA{\textsuperscript{2}TU Wien, Vienna, Austria}
\IEEEauthorblockA{\textsuperscript{3}Jagiellonian University, Krakow, Poland}
\IEEEauthorblockA{\textsuperscript{4}AGH University, Krakow, Poland}
}


\maketitle

\begin{abstract}
We present a hierarchical language-driven framework for robotic task and motion planning to improve natural, intuitive human–robot interaction in service and assistance scenarios.
The proposed system employs two large language model (LLM) modules: a high-level planning agent and a low-level spatial reasoning sub-module. 
The primary agent processes natural language commands and generates action sequences using a ReAct-style prompt, interacting with tools for object perception and manipulation (e.g., pick, place, release). For precise spatial placement, such as interpreting “place the mug next to the plate”, a separate sub-prompting module handles 3D reasoning based on object geometry and scene layout. 
The system integrates YOLOX-GDRNet for object detection and pose estimation, along with a motion execution stub. 
We evaluated the system in 24 test scenarios, ranging from simple spatial commands to high-level instructions and infeasible requests. The system achieved an overall task success rate of 86\%.
\end{abstract}


\section{Introduction}


\begin{figure}[htbp] 
    \centering
    \includegraphics[width=0.75\columnwidth, trim=0 0 11cm 0]{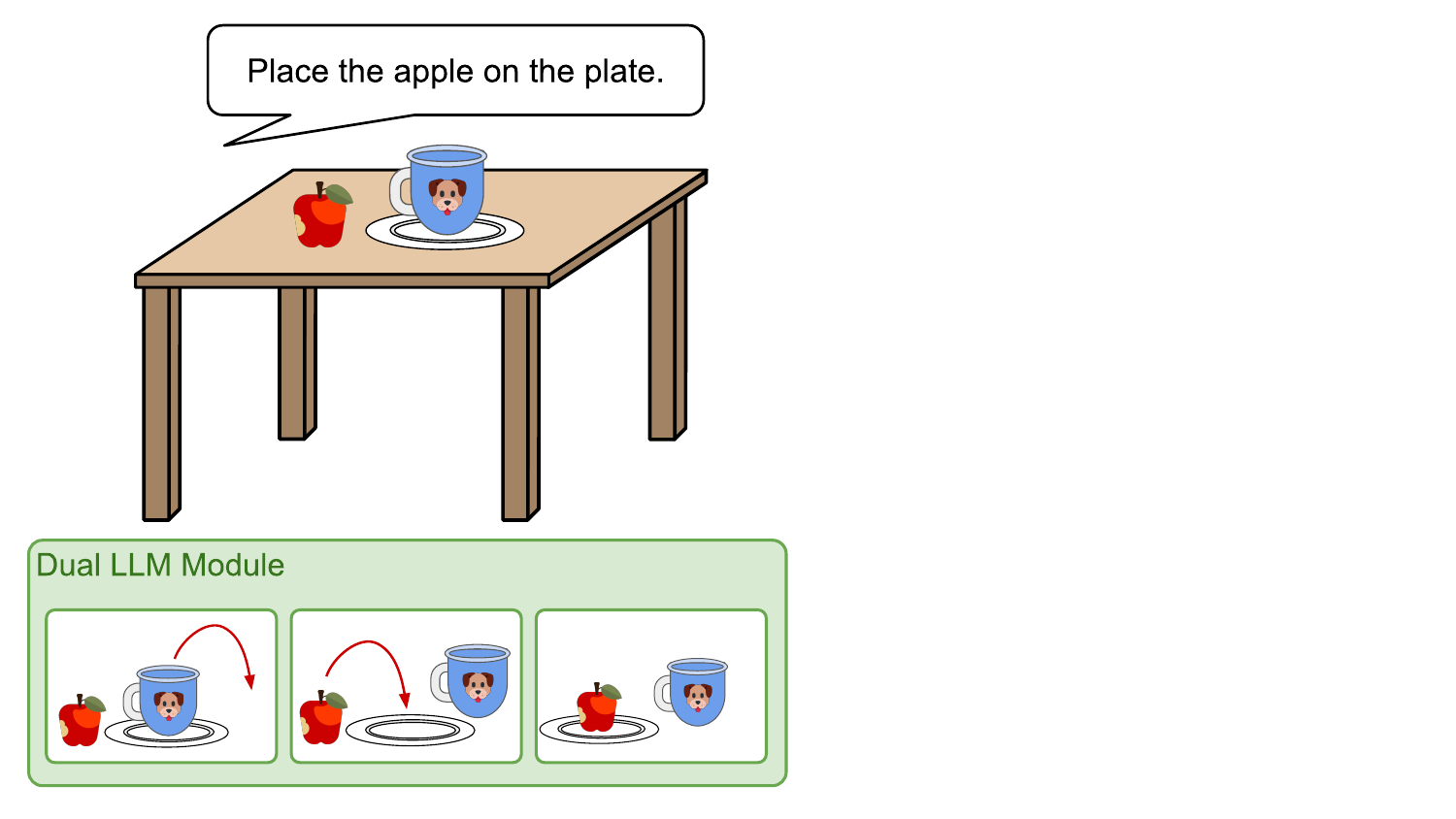}
    \caption{Overview of the Dual LLM Planer: The module reasons how to fulfill the task even when sub-tasks are required to reach the end-goal}
    \label{fig:overview}
\end{figure}

\noindent General service robots designed for home assistance must prioritize human-centered interaction to be viable in domestic settings~\cite{habibian2025survey}. 
For robots to be truly integrated in home environments, they should operate without requiring users to master specialized commands or complex interfaces~\cite{tellex2020robots}.
Instead, robots should interpret high-level natural instructions, even when those instructions are imprecise, incomplete, or based on assumptions about shared knowledge~\cite{misra2016tell}. 
Although many robotic systems today can parse natural language~\cite{zawalski2024robotic, mu2024embodiedgpt}, the transition from linguistic understanding to physically grounded action remains a major challenge.

\noindent Human language often contains ambiguities, shortcuts, and context-dependent assumptions. 
Consequently, the level of detail that a robot must infer can vary significantly from command to command. 
For example, the instruction “Pick up an apple and put it in a bowl” involves a clear location of the object and a single motor trajectory. 
A command like “Clear the table of all fruits” requires high-level semantic classification, multi-object tracking, and complex sequential planning. 
This disparity highlights that different linguistic inputs require varying degrees of reasoning, perception, and long-term planning~\cite{shridhar2020alfred}. Although smaller language models or rule-based semantic parsers can handle well-defined commands, they typically struggle with such open-ended, ambiguous instructions due to limited world knowledge and weaker compositional reasoning~\cite{petruzzellis2024assessing}.

\noindent Recent advances in Large Language Models (LLMs) provide a promising bridge for this gap~\cite{vemprala2023chatgpt}. 
LLMs demonstrate a remarkable ability to extract intent from natural language and exhibit emergent reasoning regarding world states and spatial relationships.
However, relying on LLMs for end-to-end robotic planning presents several challenges. 
Standard LLMs are not inherently grounded in physical laws and can generate ``hallucinated" plans that are kinematically or geometrically infeasible~\cite{ahn2022can}. 
Moreover, because LLMs cannot perceive the physical world directly, they rely entirely on external modules to translate raw sensory data into symbolic text.
In this paper, we propose a hierarchical agent-based framework to address these limitations. 
Our key contribution is a dual-module prompting architecture that decouples high-level task logic from low-level spatial reasoning. 
While a primary LLM-based agent manages the overall plan using a ReAct-style loop~\cite{yao2022react}, a specialized sub-module handles 3D spatial constraints and geometric feasibility. 
This modularity allows us to overcome the inconsistencies of monolithic prompting, where a single comprehensive prompt is utilized to handle all cognitive demands of a task simultaneously.
We validate our framework in a simulation environment and conduct real-world experiments, demonstrating that specialized hierarchical reasoning significantly enhances a robot's ability to execute complex human commands in unstructured environments.

The primary contributions of this work are as follows:
\begin{itemize}
    \item We introduce a dual-module architecture that decouples high-level task planning from low-level spatial reasoning to improve the reliability of generated robotic plans.
    \item We propose a specialized sub-module designed to handle 3D spatial constraints and geometric feasibility, addressing the lack of physical grounding in standard LLMs.
    \item We demonstrate the efficacy of our framework in 24 scenarios using a real-world vision-based pipeline, showing significant improvements in the execution of complex human commands in unstructured environments.
\end{itemize}

The implementation of the proposed system is publicly available online\footnote{\url{https://github.com/ichores-research/goal\_state\_reasoning}}. 




\section{Related Work}
\label{sec:related_works}

\subsection{LLMs for Task and Motion Planning}

\noindent Traditional Task and Motion Planning (TAMP) often relies on manually defined symbolic rules and PDDL (Planning Domain Definition Language)-based solvers \cite{garrett2021integrated}.
Although effective in closed environments, these methods struggle with the open-vocabulary nature of human environments.
Recent works have integrated LLMs to leverage their zero-shot reasoning for task decomposition and navigation \cite{dorbala2023can, shah2023lm}. 
These models can infer object relationships and sequences that mimic human intuition \cite{zeng2023large}.
Despite their logical reasoning abilities, these models often lack common-sense understanding.
Many LLM-based planners treat spatial coordinates as mere text, leading to commands that are logically sound but physically impossible due to object geometry or workspace constraints.
Consequently, there is a need for architectures that do not just treat LLMs as high-level schedulers, but integrate them into the geometric reasoning process to ensure physical feasibility.

\subsection{Prompting Strategies for Robotics}
\noindent The efficacy of LLMs in robotics is heavily dependent on prompt engineering. 
Chain-of-Thought (CoT) prompting has been used to break down manipulation tasks into step-by-step logic~\cite{wei2022chain, cobbe2021training}.
More advanced frameworks, such as ReAct~\cite{yao2022react}, interleave reasoning with environmental feedback, enabling dynamic re-planning.
Other approaches use LLMs to generate executable code directly from libraries~\cite{liang2023code}.
Most current prompting strategies are monolithic; they ask a single model instance to handle both the abstract logic (e.g., "What is a fruit?") and the precise spatial math (e.g., "Where is the coordinate (x,y,z) relative to the plate?"). This often results in "token exhaustion" or confused reasoning, where the model prioritizes the task logic over spatial accuracy.
We aim to fill the gap in hierarchical prompt specialization, specifically by isolating spatial geometry reasoning into a dedicated sub-module to prevent the high-level planner from becoming overwhelmed by low-level coordinate math.

\subsection{LLM-Based Planning for Household Robotics}
\noindent The application of LLMs to household robotics has been accelerated by benchmarks such as ALFRED~\cite{shridhar2020alfred} and planning frameworks including Retrieval-Augmented Planning (RAP)~\cite{kagaya2024rap} and multi-agent approaches~\cite{shi2024opex, kim2023context, bhambri2023multi}. 
While these systems excel at generating long-horizon action sequences from language and vision, transferring abstract plans to real robotic execution remains challenging. 
This gap is exacerbated by unreliable perception and limited manipulation capabilities, often restricted to basic pick-and-place actions~\cite{vanc2024tell}. 
Our work addresses this perception--action gap by leveraging LLM reasoning within a simplified but physically grounded action space. 
Unlike ALFRED-style planners, which rely on simulated affordances and simplified physics~\cite{kim2024realfred}, our approach prioritizes execution reliability on real robots, trading action expressivity for feasibility.

\section{Methodology}
\label{sec:methods}
\noindent Our framework enables a service robot to translate natural language commands into physical actions within a tabletop environment.
To ensure reliable manipulation in human-shared spaces, grasping points are pre-annotated, allowing the system to focus on the reasoning and placement logic required for assistance tasks.


\begin{figure}[htbp] 
    \centering
    \includegraphics[width=\columnwidth{}]{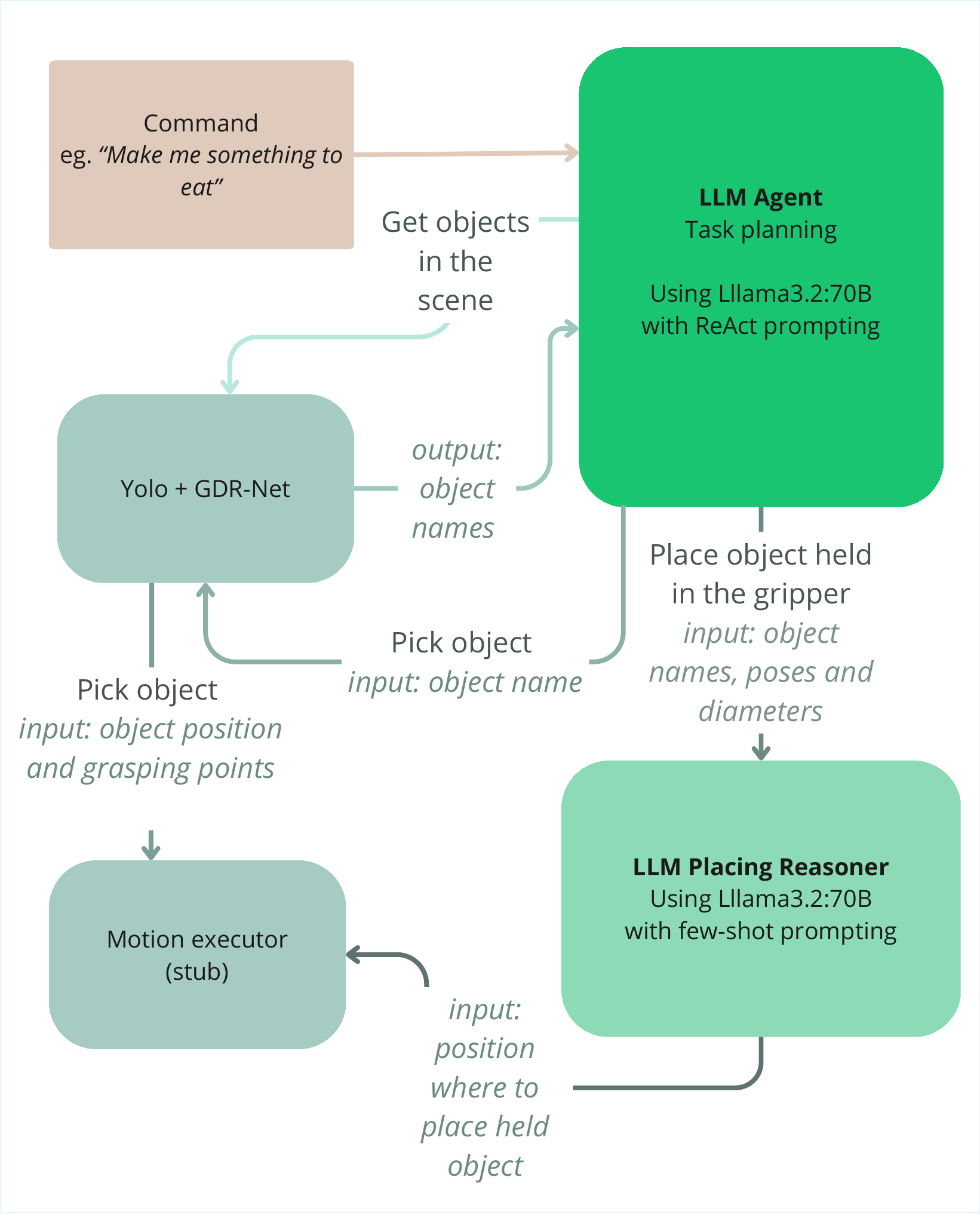}
    \caption{The system architecture integrates a YOLOX-GDRNet vision module for object detection and pose estimation, a LangChain LLM agent (powered by Llama3.2:70B with ReAct prompting) for task planning, a separate LLM placing reasoner for determining placement coordinates, and a stub simulating pick and place calls.}
    \label{fig:system_architecture}
\end{figure}

\subsection{Vision Module}

\noindent Given an input image $X \in \mathbb{R}^{2}$, a YOLOX object detector \cite{ge2021yolox} is used to obtain 2D bounding boxes and class labels for objects within the scene. 
The Region of Interest (ROI) defined by these bounding boxes is subsequently processed by a GDR-Net neural network \cite{wang2021gdr} to predict the 6D pose. 
This 6D pose is described by a 3D translation vector $\mathbf{t} \in \mathbb{R}^{3}$ and orientation defined by quaternions $\mathbf{q} \in \mathbb{H}$ within the camera coordinate frame.
All objects detected by YOLO have the corresponding 3D models stored with pre-calculated diameters $d \in \mathbb{R}$. 



\subsection{LLM Agent Planner}

\begin{figure*}[htbp]
\centerline{\includegraphics[width=\textwidth]{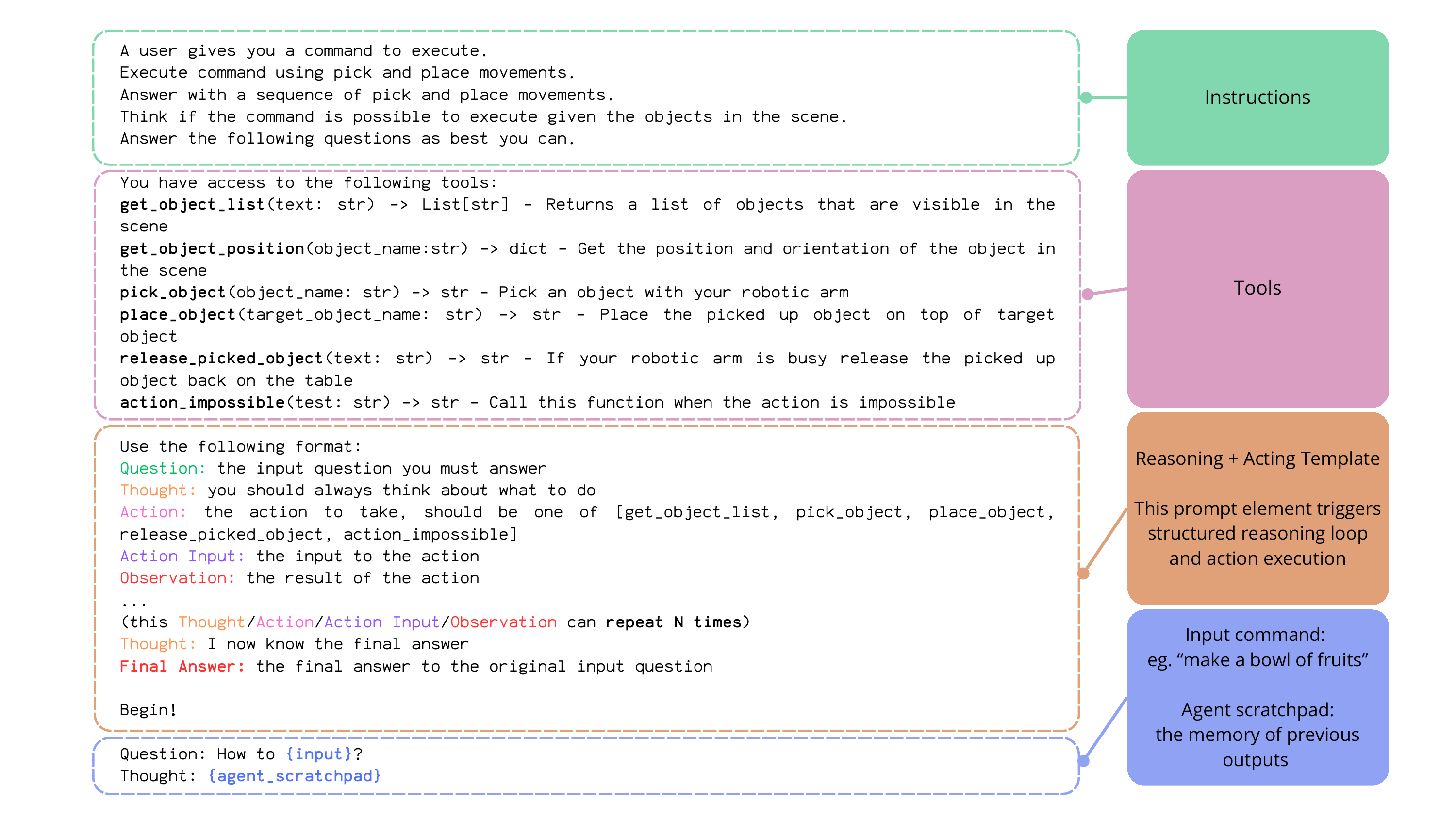}}
\caption{{The adjusted ReAct template prompt provides instructions, a list of available tools with their descriptions and signatures, and specifies the \textit{Thought-Action-Observation} format for guiding the agent's reasoning and action execution within a task.}}
\label{fig:llm_agent_prompt}
\end{figure*}

\noindent The LLM task planning agent used in this work is implemented using the Python LangChain framework \cite{noauthor_agents_nodate} and takes advantage of the capabilities of the Llama3.2:70B large language model \cite{llama3modelcard}. 
The LLM agent's operation is governed by a ReAct template prompt \cite{yao2023react}, which primes the language model to engage in an iterative reasoning and acting loop. 
This paradigm follows a structured scheme of \textit{Thought, Action, Observation} culminating in a \textit{Final Thought} that signifies the agent's completion of the current sub-task. Crucially, the agent maintains a memory of the interaction history, starting from the initial prompt. 
For subsequent steps within a longer task, the agent is prompted again, incorporating the history of its previous thoughts, actions, and the resulting observations. This self-referential prompting allows the agent to maintain context, build on prior steps, and adapt its subsequent reasoning and actions based on the outcomes of its interactions with the environment and the execution of its tools. The complete agent prompt can be found in Fig.~\ref{fig:llm_agent_prompt}. 

\noindent The agent is equipped with a suite of tools that enable it to interact with the environment and perform manipulation tasks. 
The available tools include: getting a list of objects in the scene, placing an object, picking an object, and releasing a held object in any available free space on the tabletop. 

\subsection{LLM Placing Reasoner}
\noindent A sub-prompting approach was utilized to handle tasks involving spatial reasoning, specifically for placing and releasing objects. 
This sub-prompting strategy is intended to accurately determine three-dimensional coordinates for object manipulation, taking into account the list of available objects, their locations, and their diameters. The method follows a few-shot prompting framework similar to that introduced by Wang et al. \cite{wang2023prompt}.
The main LLM agent remains unaware of these sub-prompts, preserving a modular and decoupled system architecture.
The effectiveness of the placing reasoner comes from constraining the context and input space of the model rather than relying on unconstrained language understanding. Unlike the high-level planner, this module operates on a structured representation consisting of object names, poses, and geometric properties (e.g., diameters), along with a small number of few-shot examples. This focused context reduces ambiguity and encourages the model to map spatial language (e.g., “to the left of”) to consistent geometric transformations. 

Effectively, this setup enables the breakdown of complex, multi-step manipulation tasks into smaller, more manageable components, with the language model serving as both a coordinator and a spatial reasoning module. 

\section{Evaluation \& Results}
\label{sec:result}

\subsection{Plan Generation Analysis}
\label{subsec:plan-generation}

\noindent The planner was evaluated in 24 scenarios: 16 simple tasks, 5 high-level commands, and 3 designed to be impossible.
The complexity of the scene varied, with 1 to 5 objects per scene. 
High-level commands (e.g. \textit{Make a bowl of fruits}, \textit{prepare a breakfast}) were chosen to demonstrate task diversity, while low-level commands followed the template: \textit{Place the [object] [relative location] [target]}. 
Performance was quantified by plan length (number of steps) and success rate, determined by human evaluation.
For infeasible scenarios, a task was deemed successful if the system correctly identified it as impossible.
However, we compute the number of steps needed to determine the infeasibility. 
In these cases, we measured the number of steps required by the agent to complete the infeasibility of the task.

One step in the plan is a set of \textit{Thought}, \textit{Action}, \textit{Action Input}, and \textit{Observation}.

\begin{planbox}
    \scriptsize
    \raggedright
    \noindent\textbf{\textcolor{qgreen}{Question: \textbf{How to place the banana on the left of the plate?}}} 
    
    \vspace{0.5em}\hrule\vspace{1em}

    \thought{I need to check if both the banana and the plate are present in the scene.}
    \action{get\_object\_list}
    \inputval{Check for banana and plate}
    \observation{[`029\_plate', `011\_banana']}

    \thought{Both the banana and the plate are present in the scene. I can proceed to pick up the banana.}
    \action{pick\_object}
    \inputval{011\_banana}
    \observation{You have picked up 011\_banana}

    \thought{I have successfully picked up the banana. Now I need to place it to the left of the plate.}
    \action{place\_object}
    \inputval{to the left of the 029\_plate}
    \observation{You have placed 011\_banana to the left of the 029\_plate}

    \thought{I now know the final answer.}

    \vspace{0.5em}\hrule\vspace{1em}
    
    \noindent\textbf{\textcolor{fanswer}{Final Answer: The banana has been successfully placed to the left of the plate.}}
\end{planbox}

\noindent As summarized in Table~\ref{tab:evaluation_results}, the system was evaluated in 24 test scenarios.
The system achieved an overall success rate of 86\% with an average planning duration of 5 steps per task.
For simple commands (16 scenarios), the agent averaged 4 steps.
This complexity increased for high-level commands (5 scenarios), which required 6 steps on average. In infeasible scenarios (3 scenarios), the system achieved a 100\% success rate \, correctly identifying the task's impossibility with an average of 5 steps before termination.

\begin{table}[h]
\centering
\caption{Performance Summary Across Scenario Categories}
\label{tab:evaluation_results}
\begin{tabular}{lccc}
\toprule
\textbf{Scenario Category} & \textbf{Count} & \textbf{Success Rate} & \textbf{Avg. Steps} \\ \midrule
Simple Commands            & 16             & 87.5\%                & 4                   \\ \addlinespace[0.5em]
High-Level Commands        & 5              & 80.0\%                & 6                   \\ \addlinespace[0.5em]
Infeasible Scenarios       & 3              & 100.0\%               & 5                   \\ \midrule
\textbf{Overall}           & \textbf{24}    & \textbf{86.0\%}       & \textbf{5}          \\ \bottomrule
\end{tabular}
\end{table}

\subsection{Qualitative Evaluation of the End Scene}
\begin{figure}
    \centering
    \includegraphics[width=\columnwidth{}]{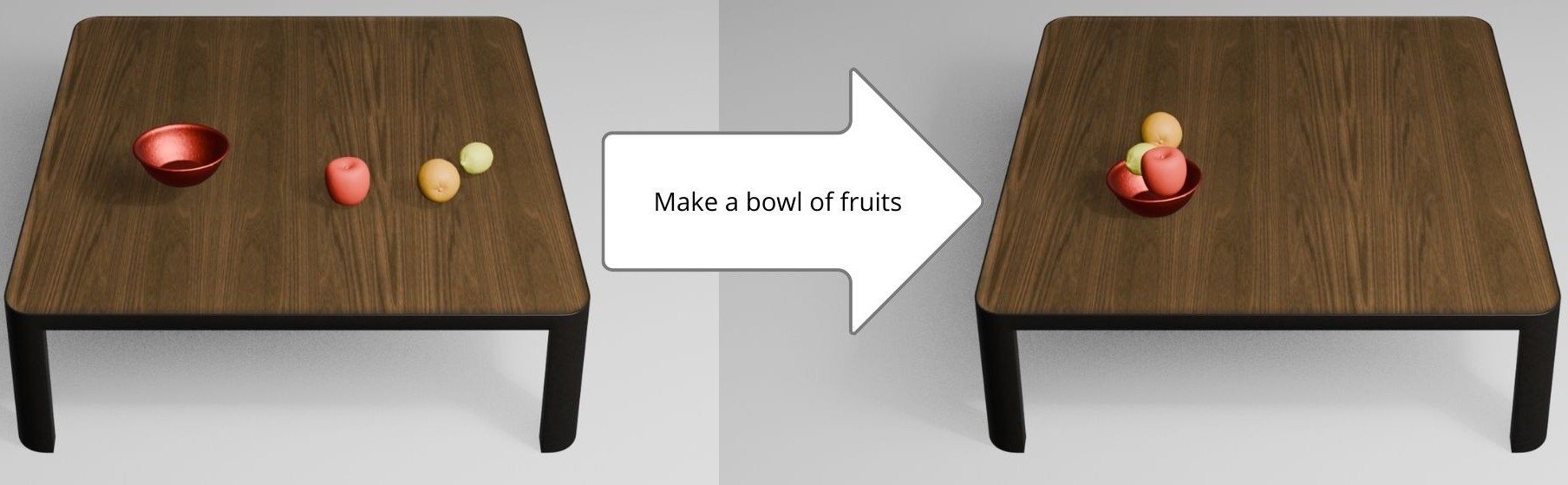}
    \caption{Example of a high-level scenario (left). The resulting scene is evaluated by humans on a scale from 1 to 10.}
    \label{fig:end_scene}
\end{figure}

\noindent We perform a qualitative evaluation based on human judgment to assess the success of the task in 24 scenarios. Three independent evaluators rated the alignment between initial commands and final scenes on a scale of 1 to 10, focusing on semantic and visual plausibility. 
This subjective metric allows for a more nuanced assessment of success, particularly for ambiguous spatial instructions, ensuring that the model's performance aligns with human expectations for intuitive service robotics.

\noindent For this evaluation, we select seven representative cases: five high-level commands, one simple command involving an ambiguous \textit{"near"} location specifier, and an infeasible command.
The starting scenes were visualized using Blender4.4\footnote{https://www.blender.org/download/releases/4-4/}, and the final scenes were rendered based on the LLM-generated plans (see Fig.~\ref{fig:end_scene}).

\noindent The mean rating across all scenes was 6.86, indicating a generally high performance with significant variation between tasks (ranging from 3.0 to 10.0).
This variance reflects the inherent complexity of spatial and semantic requirements.

\noindent To assess the degree to which the annotator is agreeable, we calculated the standard deviation ($\sigma$) of the ratings for each scene. 
Tasks with clear spatial relationships (e.g., ``Make a bowl of fruits") exhibited low variance, whereas more ambiguous tasks (e.g., ``Make a salty snack'') resulted in greater disagreement. 
The average standard deviation in all tasks was approximately 1.66, suggesting moderate consistency among judges. Table~\ref{tab:eval_results} summarizes the individual scene scores and the corresponding standard deviations.



\begin{table}[htbp]
\centering
\caption{Summary of human evaluation scores for final scenes}
\label{tab:eval_results}
\begin{tabular}{lcc}
\toprule
\textbf{Scene} & \textbf{Mean Rating} & \boldmath$\sigma$ \\ \midrule
Make a bowl of fruits                                 & 9.67  & 0.57 \\ \addlinespace[0.3em]
Create citrus display                                 & 8.67  & 1.15 \\ \addlinespace[0.3em]
Place the mug near the fruits                         & 8.00  & 2.00 \\ \addlinespace[0.3em]
Prepare breakfast                                     & 5.67  & 2.30 \\ \addlinespace[0.3em]
Stack up the dishes                                   & 3.00  & 2.00 \\ \addlinespace[0.3em]
Make a salty snack                                    & 3.00  & 2.64 \\ \addlinespace[0.3em]
\makecell[l]{Place the mustard in front of \\ the dishes (impossible)} & 10.00 & 0.00 \\ \bottomrule
\end{tabular}
\end{table}

\noindent In general, these results show that while the model usually predicts plausible outcomes, its success depends largely on how evaluators interpret the task instructions. 
Furthermore, human judges tend to favor visual appearance over strict rules, especially when the final scene appears natural.

\subsection{Spatial Understanding Evaluation}
\noindent To evaluate the spatial reasoning capabilities of the placing reasoner, we designed 16 test scenarios requiring the model to position an object relative to a target (or group) using spatial specifiers such as \textit{to the left of}, \textit{on top of}, and \textit{near}.
Performance is assessed by analyzing the distribution of predicted coordinates (x, y, z) for each term.
By aggregating predictions across multiple scenes, we characterize the model's geometric interpretation of spatial language.

\begin{figure}[htbp]
    \centering
    \includegraphics[width=\columnwidth{}]{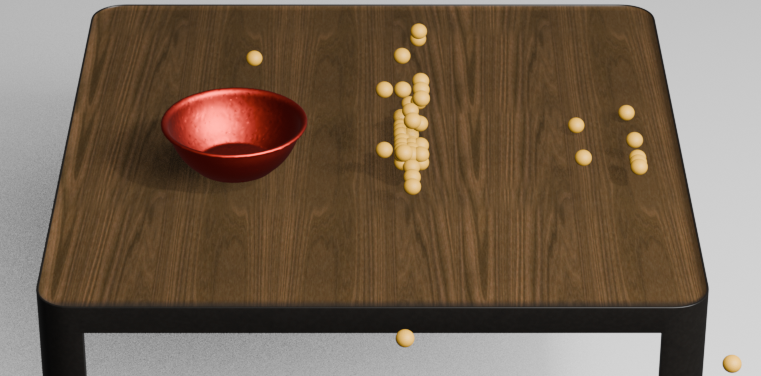}
    \caption{Visualization of outputs from vision systems combined with LLM placing reasoner. The orange spheres illustrate where an orange should be placed in response to a command: \textit{Place the orange to the right of the bowl}. }
    \label{fig:to_the_right}
\end{figure}

\noindent Specifically, we calculate the mean offset of the placed object relative to the target's centroid; for instance, commands involving \textit{to the left of} the axis are expected to produce a distinct offset along the x-axis.
To evaluate reliability, we measure the variance in the predicted positions in repeated trials.
Low variance indicates a stable and coherent internal representation of spatial relationships.
The results of this evaluation are summarized in Table~\ref{tab:performance_by_location_styled_format}.

\begin{table}[htbp]
\centering
\caption{Coordinate Distribution and Variance by Location Specifier}
\label{tab:performance_by_location_styled_format}
\begin{tabular}{lcccc}
\toprule
\textbf{Location} & \textbf{Avg. Variance ($m^2$)} & \boldmath$\Delta X$ & \boldmath$\Delta Y$ & \boldmath$\Delta Z$ \\ \midrule
on top            & 0.00                   & 0.00                & 0.00                & -0.26               \\ \addlinespace[0.4em]
next to           & 0.00                   & -0.07               & 0.00                & 0.02                \\ \addlinespace[0.4em]
left              & 0.01                   & 0.34                & 0.03                & 0.04                \\ \addlinespace[0.4em]
right             & 0.02                   & -0.19               & 0.00                & 0.01                \\ \addlinespace[0.4em]
near              & 0.02                   & 0.01                & 0.02                & -0.04               \\ \addlinespace[0.4em]
inside            & 0.03                   & -0.03               & 0.01                & 0.03                \\ \addlinespace[0.4em]
in front          & 0.08                   & -0.02               & -0.00               & 0.07                \\ \addlinespace[0.4em]
behind            & 0.76                   & -0.17               & 0.07                & -0.00               \\ \bottomrule
\end{tabular}
\end{table}

\noindent The \textit{on top} and \textit{next to} relations demonstrate the highest precision, characterized by the lowest average coordinate variance.
The variance in the reasoner's output increases for \textit{left}, \textit{right}, \textit{near}, and \textit{inside}, with \textit{behind} showing the greatest variability. 
Since the model temperature was set to 0, the observed variance is primarily attributable to fluctuations in the 2D bounding boxes and 6D pose estimations rather than stochasticity in the LLM's text generation.
The \textit{left} and \textit{right} specifiers show the most significant offsets along the X-axis.
The \textit{on top} location is distinct in that its offset is almost exclusively confined to the Z-axis.
Ultimately, the spatial reasoning module shows varying degrees of proficiency. 
While it excels at localizing objects using \textit{on top} and \textit{next to}, while it effectively distinguishes between \textit{left} and \textit{right}, its performance is less consistent when interpreting the depth-based relations \textit{in front} and \textit{behind}.

\subsection{Physical Robot Evaluation}

\begin{figure}[htbp]
\centering
\includegraphics[width=0.75\columnwidth]{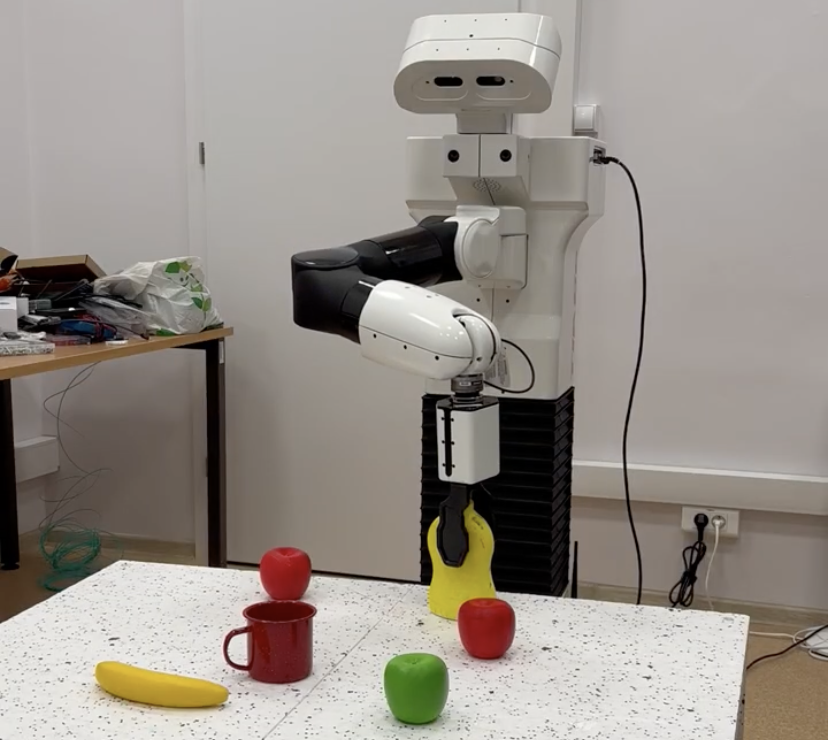}
\caption{The real-world experimental setup featuring the PAL Robotics Tiago platform executing a command to retrieve a mustard bottle from a cluttered surface.}
\label{fig:tiago}
\end{figure}

To validate the proposed system in a physical setting, we conducted experiments on a PAL Robotics Tiago manipulator equipped with a parallel gripper (Fig. \ref{fig:tiago}). The setup mirrored vision-only simulation experiments, with the passive motion stub replaced by an active control interface (Fig. \ref{fig:system_architecture}) that executes real trajectories and reports execution outcomes. The robot was positioned in front of a table that contained target objects and distractors.

The evaluation focused on grasping and reasoning at the task-level. Although the vision and semantic reasoning modules behaved consistently with simulation results (Section \ref{subsec:plan-generation}), the real-world testing was limited to \emph{pick} actions due to hardware integration constraints; consequently, the placement reasoning module was not evaluated.

The system reliably identified target objects and inferred user intent in all trials, despite clutter. We evaluated the grasps of a mustard bottle, an apple, and a can of spam under varying clutter conditions, with the reasoning system producing correct plans in all cases. Observed failures were confined to low-level execution, specifically motion planning (MoveIt) and extrinsic calibration errors. For example, grasping the spam can require minor manual correction due to camera miscalibration, and the apple trial involved a minor collision with a distractor. Importantly, these failures did not originate from semantic or task-level reasoning.





\section{Conclusions}

\noindent Our results demonstrate that LLMs effectively ground natural language in 3D geometric relations, exhibiting a robust spatial grammar suitable for zero-shot placement. 
However, real-world utility remains constrained by limitations in physical feasibility and environmental generalization. 
Future research should integrate geometric priors and physical simulators to refine predictions within dynamic, embodied settings. 
Ultimately, merging the symbolic flexibility of LLMs with grounded spatial perception offers a promising path toward intuitive and generalizable human-robot collaboration.

\section*{Acknowledgements}

This work was supported in part by the National Science Center, Poland, under the OPUS call within the Weave program (project no. 2021/43/I/ST6/02489), and in part by funds allocated to AGH University by the Polish Ministry of Education and Science.

\bibliographystyle{IEEEtran}
\bibliography{example}

\end{document}